\pdfoutput=1
\documentclass[11pt]{article}

\usepackage[preprint]{acl}

\usepackage{times}
\usepackage{latexsym}

\usepackage[T1]{fontenc}
\usepackage[utf8]{inputenc}

\usepackage{microtype}

\usepackage{inconsolata}

\usepackage{graphicx}

\usepackage{times}
\usepackage{epsfig}
\usepackage{amsmath}
\usepackage{amssymb}
\usepackage{float}
\usepackage{placeins}
\usepackage{multirow}
\usepackage{breakurl}
\usepackage{comment}
\usepackage{hyperref}  % To ensure clickable links
\title{ControlText: Unlocking Controllable Fonts in Multilingual Text \\ Rendering without Font Annotations}

\author{
  Bowen Jiang\textsuperscript{1}\ ,
  Yuan Yuan\textsuperscript{1}\footnotemark[1]\ ,
  Xinyi Bai\textsuperscript{2},
  Zhuoqun Hao\textsuperscript{1},
  Alyson Yin\textsuperscript{3}, \\
  \textbf{Yaojie Hu\textsuperscript{1},
  Wenyu Liao\textsuperscript{1},
  Lyle Ungar\textsuperscript{1},
  Camillo J. Taylor\textsuperscript{1}}\\
  \begin{tabular}{ccc}
    University of Pennsylvania\textsuperscript{1} & 
    Cornell University\textsuperscript{2} &
    University of California, Irvine\textsuperscript{3} \\
    Philadelphia, PA 19104 & Ithaca, NY 14850 & Irvine, CA 92697
  \end{tabular} \\
 \small{
    \href{mailto:bwjiang@seas.upenn.edu, yyuan86@seas.upenn.edu, cjtaylor@seas.upenn.edu}{{bwjiang, yyuan86}@seas.upenn.edu}, 
    \href{mailto:ungar@cis.upenn.edu}{ungar@cis.upenn.edu}, 
    \href{mailto:cjtaylor@seas.upenn.edu}{cjtaylor@seas.upenn.edu}
}
}

\begin{document}
\maketitle

\begin{figure*}[!t]
    \centering
    \includegraphics[width=0.84\linewidth]{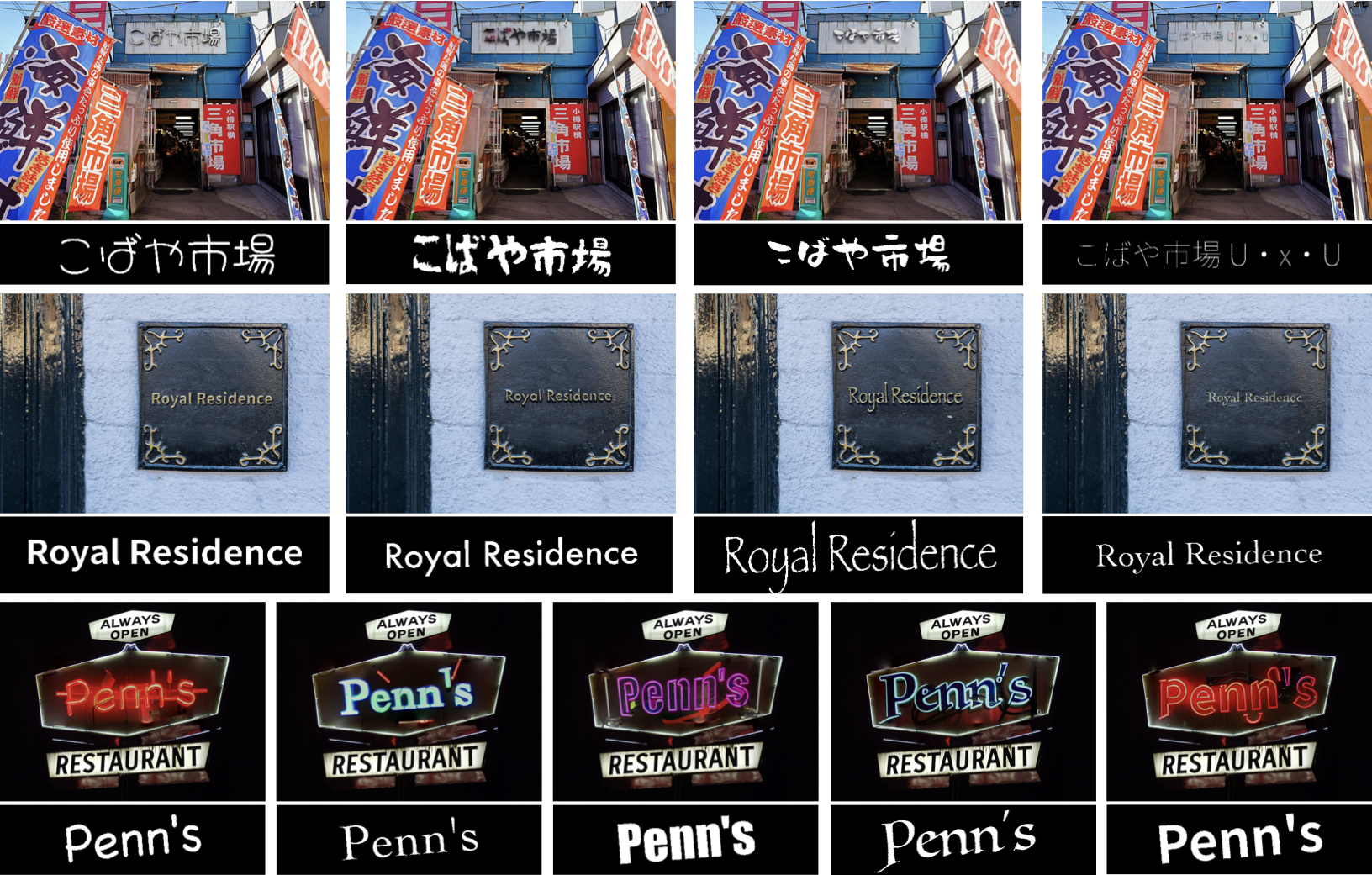}
    \caption{Examples of real-world test images with text generated by ControlText in various fonts and languages. Each row presents both the rendered images and the textual part of the corresponding glyph controls that provide the text and the intricate font information in pixel space.}
    \label{fig:top}
\end{figure*}

%%%%%%%%% BODY TEXT
%%%%%%%%% ABSTRACT
\begin{abstract}
   This work demonstrates that diffusion models can achieve \textit{font-controllable} multilingual text rendering using just raw images without font label annotations.
   Visual text rendering remains a significant challenge. While recent methods condition diffusion on glyphs, it is impossible to retrieve exact font annotations from large-scale, real-world datasets, which prevents user-specified font control.
   To address this, we propose a data-driven solution that integrates the conditional diffusion model with a text segmentation model, utilizing segmentation masks to capture and represent fonts \emph{in pixel space} in a \textit{self-supervised} manner, thereby eliminating the need for any ground-truth labels and enabling users to customize text rendering with any multilingual font of their choice. The experiment provides a proof of concept of our algorithm in zero-shot text and font editing across diverse fonts and languages, providing valuable insights for the community and industry toward achieving generalized visual text rendering. Code is available at \href{https://github.com/bowen-upenn/ControlText}{github.com/bowen-upenn/ControlText}
   
   % This work demonstrates that diffusion models can achieve font-controllable multilingual text rendering using just raw images in the open world without font annotations. Visual text rendering remains a significant challenge for diffusion models due to the intricate structure of textual strokes they must learn from data. While recent works use diffusion models conditioned on glyphs to guide the generation process, it is impossible to retrieve exact font annotations from large-scale, real-world datasets, which prevents these algorithms from allowing users to specify fonts for the rendered texts. To address this, we propose a data-driven solution that integrates the conditional diffusion model with a text segmentation model, utilizing segmentation masks to capture and represent fonts \emph{in pixel space}  in an unsupervised manner, thereby eliminating the need for any ground-truth labels. A text recognition model also helps validate the fidelity of these conditional maps, ensuring accurate text representation. We present this algorithm as part of a user-in-the-loop copilot tool for real-world design tasks, supporting existing and user-designed fonts across multiple languages. The experiment provides a proof of concept of our algorithm in zero shot visual text generation and editing across diverse fonts and languages, providing valuable insights for the community and industry toward achieving generalized visual text rendering in images. Code will be made available at \href{https://github.com/bowen-upenn/ControlText}{github.com/bowen-upenn/ControlText}.
\end{abstract}
\section{Introduction}
Diffusion model is one of the dominant paradigms in image generation~\cite{ho2020denoising, saharia2022photorealistic, zhang2023adding, rombach2022high, betker2023improving, ramesh2022hierarchical, esser2024scaling}, because of its iterative denoising process that allows fine-grained image synthesis. While these models effectively capture data distributions of photorealistic or artistic images, they still fall short in generating high-fidelity text. Rendering text in images is inherently more challenging as it requires precise knowledge of the geometric alignment among strokes, the arrangement of letters as words, the legibility across varying fonts, sizes, and styles, and the integration of text into visual backgrounds. At the same time, humans are more sensitive to minor errors in text, such as a missing character or an incorrectly shaped letter, compared to natural elements in a visual scene that allow for a much higher degree of variation.

Increasing attention has been paid to visual text rendering~\cite{bai2024intelligent, han2024ace, li2024hfh} due to its high user demands. Instead of relying solely on diffusion models to remember exactly how to render text, recent research is starting to embed the visual attributes of texts, such as glyphs~\cite{tuo2023anytext, liu2024glyph, ma2024glyphdraw2, yang2024glyphcontrol}, as input conditions to diffusion models. However, it is still difficult for users to specify the desired font in the open world, and there remain open challenges that burden the development of font-controllable text rendering:
% \vspace{-1mm}
\begin{itemize}
\item No ground-truth font label annotation is available in the massive training dataset, while synthetic images often fail to accurately mimic subtle details that appear in reality.
% \vspace{-1mm}
\item There are numerous fonts available in the open world, but many fonts with different names are very similar, which confounds evaluation.
% \vspace{-2mm}
\item Users like visual designers may want to explore different fonts during their design process, even creating novel fonts of their own. 
% In addition, the user-selected layout locations may not be exactly what is intended.
% \vspace{-1mm}
\end{itemize}

This work aims to address the above challenges. To summarize our contributions, we introduce the simplest and, to our best knowledge, one of the few~\cite{ma2024glyphdraw2, liu2024glyph} open-source methods for rendering visual text with user-controllable fonts. 
We provide code in the hope that others can draw inspiration from the underlying \textbf{data-driven algorithm} and benefit from the \textbf{simplicity in the self-supervised training}.

We also provide the community with a comprehensive dataset for font-aware glyph controls collected from diverse real-world images. We further propose a \textbf{quantitative evaluation metric for handling fuzzy fonts in the open world.} Experimental results demonstrate that our method, ControlText, as a text and font editing model, facilitates a human-in-the-loop process to generate multilingual text with user-controllable fonts in a zero-shot manner.

\section{Related Work}
% \subsection{Visual Text Rendering}
\paragraph{Generation from Prompts or Text Embeddings} Text-to-image generation~\cite{zhang2023text, bie2023renaissance} has advanced significantly in recent years, leveraging conditional latent diffusion models~\cite{ho2020denoising, rombach2022high, zhang2023adding}. Foundational image generation models~\cite{ramesh2021zero, betker2023improving, midjourney, saharia2022photorealistic, flux1_ai, esser2024scaling, yang2024cross, zhao2023unleashing, hoe2024interactdiffusion, sun2025anycontrol, chang2022maskgit} have achieved remarkable progress in creating high-quality photo-realistic and artistic images. 

Despite these advancements, visual text rendering~\cite{bai2024intelligent, han2024ace, li2024hfh} continues to pose significant challenges. Several algorithms rely on text embeddings from user prompts or captions to control the diffusion process, such as TextDiffuser~\cite{chen2024textdiffuser}, TextDiffuser2~\cite{chen2025textdiffuser}, and DeepFloyd's IF~\cite{deepfloyd_if}. \citet{li2024empowering} utilizes intermediate features from OCR~\cite{du2020pp} as text embeddings, \citet{liu2022character, wang2024high, choi2024towards} take one step deeper into the character level,and TextHarmony~\cite{zhao2024harmonizing} queries a fine-tuned vision-language model to generate embeddings from images and captions. 

\paragraph{Generation from Glyphs} The majority of algorithms rely on visual glyphs, pixel-level representations of texts, to guide the generation process. However, because most image training datasets lack ground-truth font annotations,  \textbf{most algorithms utilize a fixed standard font to render the texts on their glyph controls.} 
For instance, GlyphControl~\cite{yang2024glyphcontrol} and GlyphDraw~\cite{ma2023glyphdraw} render OCR-detected text using a fixed font, with the former adding a glyph-specific ControlNet~\cite{zhang2023adding}. TextMaster~\cite{wang2024textmaster}, DiffUTE~\cite{chen2024diffute}, and AnyTrans~\cite{qian2024anytrans} enforce font consistency within images, while \citet{zhang2024brush} introduces font variation through random sampling. Layout generation is also addressed using language and vision models~\cite{tuo2023anytext, zhu2024visual, li2024first, seol2025posterllama, lakhanpal2024refining, paliwal2024orientext, zhao2024ltos}. In contrast, we focus on human-in-the-loop text editing without large language or multimodal models.

Our work builds upon the codebase of AnyText~\cite{tuo2023anytext}, a glyph-based algorithm that trains a base ControlNet~\cite{zhang2023adding} model to render visual texts, with glyphs being generated in a fixed standard font due to unavailability of ground-truth font annotations, leaving the model to infer an appropriate font.

\paragraph{Font-Controllable Generation} Fewer recent works are more closely related to ours in enabling controllable fonts~\cite{tuo2024anytext2, ma2024glyphdraw2, li2024joytype, shi2024fonts, paliwal2024customtext, liu2025glyph}. \textbf{However, none of these works provide a quantitative evaluation metric to assess the generated fonts in open-world settings.} AnyText2~\cite{tuo2024anytext2} is a concurrent work developed by the authors of AnyText~\cite{tuo2023anytext}. We share a similar architecture \cite{tuo2024anytext2}, but we eliminate its use of lengthy language prompts in the inputs, separate models to support different languages, and the trainable OCR model to encode the font features. Instead, we use OCR solely to filter out low-quality glyph controls. \citet{tuo2024anytext2} is also not yet open-sourced at the time of our submission.

Several works tackle font control through predefined font labels or additional supervision. 
Glyph-ByT5~\cite{liu2025glyph, liu2024glyph} requires language-specific font labels and emphasizes language understanding, while we treat text rendering as purely visual. 
GlyphDraw2~\cite{ma2024glyphdraw2} learns font features via cross-attention and a fine-tuned language model, but lacks quantitative font evaluation. 
JoyType~\cite{li2024joytype} focuses on synthetic e-commerce images with 10 fixed fonts and vision-language models, assuming OCR is font-sensitive—unlike our font-agnostic assumption. 
FonTS~\cite{shi2024fonts} and CustomText~\cite{paliwal2024customtext} rely on pre-specified font labels or user-defined font names. 
While \citet{liu2024glyph} highlights challenges with small fonts, we show that localized editing improves small-font quality.
\textbf{In contrast to above methods, we eliminate the need for font labels, special tokens, or predefined font names} \cite{liu2024glyph, shi2024fonts, paliwal2024customtext}\textbf{, enabling generalization to unseen fonts and languages.}

\section{Technical Approach}

\begin{figure*}[!t]
    \centering
    \includegraphics[width=\linewidth]{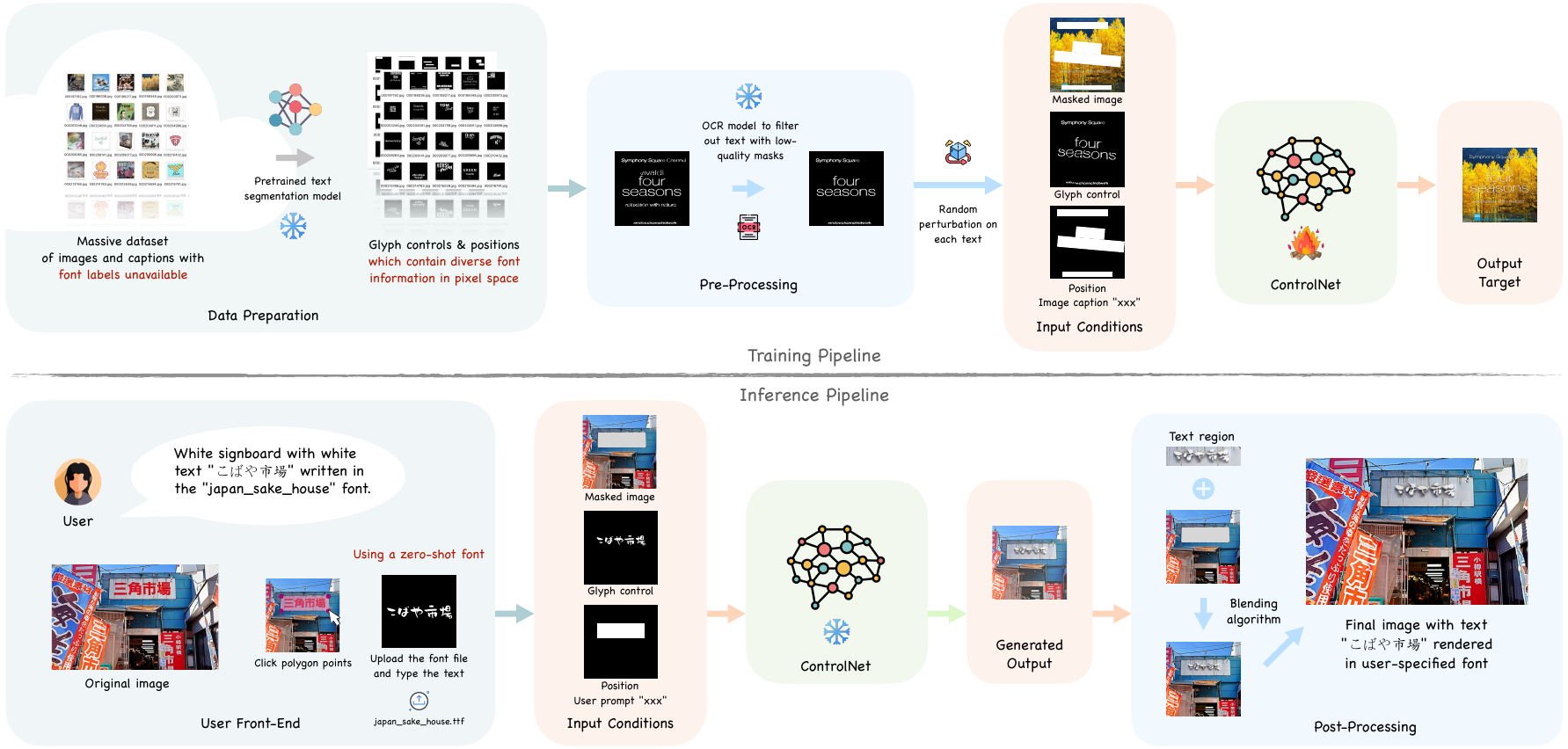}
    \caption{System overview. It consists of two parts (1) Training pipeline: text segmentation masks are extracted as glyph controls from a large image dataset without ground-truth font annotations. Low-quality masks are filtered out using an OCR model, and random perturbations are applied to prevent the model from overfitting to exact pixel locations of the glyphs. (2) Inference pipeline: users upload images, specify text regions, and provide any desired font file through the user front-end. The model generates an image patch with the rendered text, which is then seamlessly blended into the original image. Throughout this figure, models marked with a fire icon indicate trainable weights, while those marked with a snowflake icon are frozen.}
    \label{fig:flow}
\end{figure*}

\begin{figure}[t]
    \centering
    \includegraphics[width=\linewidth]{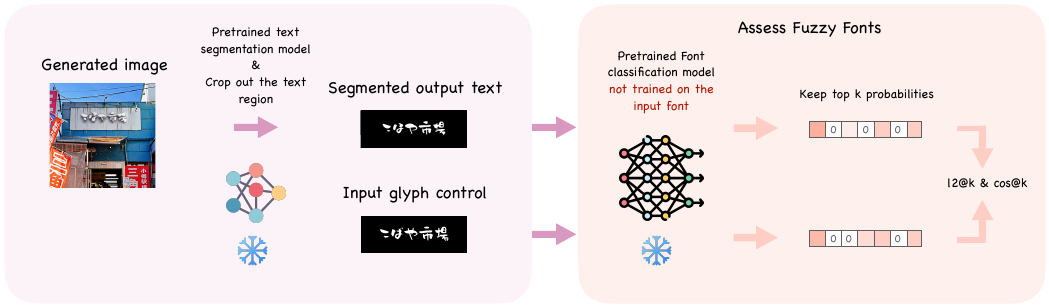}
    \caption{Evaluation pipeline: the cropped regions of the generated text and the input glyph are processed by a pretrained font classification model, which may not have seen the user-specified font. The proposed $l_2@k$ and $\cos@k$ metrics for fuzzy fonts assume that similar fonts have similar output probability vectors, while we retain only top-$k$ values while zeroing out the rest.}
    \label{fig:eval}
\end{figure}

We envision this method being used as a modular plug-in for existing text-to-image generation frameworks. It works with images generated by any base models or actual photos. For instance, when incorrect text is generated, or the user wants to replace some text or modify its font, our algorithm can be specifically targeted to these localized regions without altering remaining parts in images. By leveraging a human-in-the-loop approach, the model aims to render controllable visual text within the user-specified region, perform background inpainting, and blend the modified region back into the original image, regardless of its original size.

\subsection{Data-Driven Insights}

ControlText enables user-controllable font rendering through a simple, data-driven approach, without complex architectures, embracing the principles of the bitter lesson~\cite{sutton2019bitter}. By training on diverse unsupervised glyphs rich in pixel-level font details, the diffusion model learns to reconstruct images directly from visual cues. 
\textbf{The key insight is that the model learns to use pixel-level glyphs as direct cues for text generation, eliminating the need for font labels. Glyphs can mimic any target font, with guidance provided solely by their visual appearance.}

In inference, \textbf{the model should have seen a diverse set of glyphs during training, including intricate font features represented by pixel details near the textual edges in the glyphs.} With this information, it can render unseen languages or unfamiliar text without requiring prior knowledge of how to write the text from scratch, how to arrange individual letters or characters, or understanding their semantic meaning. \textbf{The model just treats text as a collection of pixels rather than linguistic entities.} This self-supervised data-driven approach not only enhances the model's generalizability to open-world scenarios, but also ensures scalability when more image data, computation, and larger base models become available.

\subsection{Training Pipeline}
\subsubsection{Collection of Font-Aware Glyphs} \label{sec:prepare_glyph}
Our training pipeline begins with the collection of glyph controls by performing text segmentation on images. We use TexRNet~\cite{xu2021rethinking} as our text segmentation algorithm to identify text regions and provide fine-grained masks, preserving intricate features of different fonts in pixel space, It also provides bounding boxes that will serve as position controls~\cite{tuo2023anytext}. 
% \cite{xu2021rethinking} provides masks for words, as well as word effects like shadows or stereo effects over background objects. We on-purposely exclude masks for word effects, as we want to avoid requiring users to manually sketch these effects, but rely on the diffusion models to infer and render them automatically. 
The segmentation algorithm is a pre-trained deep-learning-based model, so it may occasionally miss masks for certain letters or parts of letters. As a result, we introduce an OCR model, specifically PaddleOCR-v3~\cite{du2020pp}, into the pipeline following the segmentation process. The OCR model validates the detected text by filtering out masks that fail to meet our quality criteria, regardless of the font: (1) an OCR confidence score no lower than 0.8. (2) an edit distance no greater than 20\% of the string length. This step ensures that only high-quality segmentation masks are retained as glyph controls.

% The glyph preparation process can be expressed as
% \begin{equation}
%     \mathcal{M} (\boldsymbol{I}) = \tilde{\boldsymbol{c}}_{g} \in \mathbb{R}^{n \times n} \xrightarrow{OCR} \boldsymbol{c}_{g} \in \mathbb{R}^{n \times n},
% \end{equation}
% where $\boldsymbol{I} \in \mathbb{R}^{n \times n \times 3}$ is the target image, $\mathcal{M}$ is the text segmentation model, $\operatorname{OCR}$ means removing texts with low OCR scores in the pixel space, and $\tilde{\boldsymbol{c}}_{g}$ and $\boldsymbol{c}_{g}$ are the glyph control before and after the OCR-based filtering process.

\subsubsection{Perspective Distortion}
Segmentation masks, even after quality filtering, are not directly usable as glyph controls. In real-world scenarios, \textbf{users are unlikely to specify the exact locations of text or precisely align the text with the background}. To address this issue, we apply random perspective transformations to the collected glyph images, introducing slight translations and distortions to the text, without affecting the fonts. Specifically, we add random perturbations to the four corner points of the text's bounding box, with the perturbation upper-bounded by $\epsilon$ pixels.

\begin{comment}
Using the original corner points of the bounding box
\begin{equation}
\left\{ (x_i, y_i)\ \text{for} \ i \ \in\{1,2,3,4\} \right\}
\end{equation}
and the randomly perturbed ones 
\begin{align}
& \left\{ (x_i + \delta x_i, y_i + \delta y_i)\ \text{for} \ i \ \in\{1,2,3,4\} \right\}, \\
& \text{ s.t. }\left|\delta x_i\right| \leq \epsilon \text{ and }\left|\delta y_i\right| \leq \epsilon,
\end{align}
\end{comment}

We then compute a homography matrix $\boldsymbol{M} \in \mathbb{R}^{3 \times 3}$ that maps the original text region to a slightly distorted view. \textbf{This design ensures that the diffusion model does not rigidly replicate the exact pixel locations of the glyphs} but instead learns to adaptively position the text in a way that best integrates with the output image.

\subsubsection{Main Training Process} \label{sec:train_input}
The diffusion process builds upon AnyText~\cite{tuo2023anytext}, leveraging ControlNet~\cite{zhang2023adding} as the base model. As shown in Figure~\ref{fig:flow}, the model takes the following five inputs during training. We expect the training dataset to consist of images containing text, captions, and polygons for the text region. The text and polygons can be automatically extracted using an OCR algorithm.
\begin{itemize}  
    \item Font-aware glyph control \(\boldsymbol{c}_{g} \in \mathbb{R}^{n \times n}\): A binary mask representing the text and its font features in pixel space.
    % where textual pixels are assigned a value of 1 and background pixels are 0.  
    \item Position control \(\boldsymbol{c}_{p} \in \mathbb{R}^{n \times n}\): A binary mask for the bounding box of the text region.
    % with the bounding box pixels set to 1 and the background pixels set to 0.  
    We restrict ourselves to square local image regions. % for simplicity.
    \item Masked image \(\boldsymbol{c}_{m} \in \mathbb{R}^{n \times n \times 3}\): An RGB image normalized to the range \([-1, 1]\), where the region within the box position \(\boldsymbol{c}_{p}\) is masked to 0. Every other pixel is identical to the target image $\boldsymbol{I} \in \mathbb{R}^{n \times n \times 3}$.
    \item Image caption \(\boldsymbol{c}_{l}\): We adopt the same handling approach in~\citet{tuo2023anytext}, except for using our own $\boldsymbol{c}_{g}$. We empirically observe that image captions are not crucial for this work.
    \item Random noise input \(\boldsymbol{x}_{T} \in \mathbb{R}^{m \times m \times d}\) in the embedding space to initialize the reverse denoising process~\cite{ho2020denoising}.
\end{itemize}  
The model outputs a denoised tensor $\boldsymbol{x}_{T} \in \mathbb{R}^{m \times m \times d}$ after $T$ timesteps, which can be reconstructed back to an image $\boldsymbol{\hat{I}} \in \mathbb{R}^{n \times n \times 3}$.
% trained to approximate the ground-truth target image $\boldsymbol{I} \in \mathbb{R}^{n \times n \times 3}$.
In the expressions above, $n$ denotes the edge length of the image, $m$ represents the spatial resolution of the hidden features in the latent diffusion~\cite{ho2020denoising}, and $d$ represents the number of channels. 

We concatenate all input conditions as $\boldsymbol{c}$ and perform the following reverse denoising process:
\begin{align}
    & \boldsymbol{c} = \varphi(\operatorname{cat}[\xi_g(\boldsymbol{c}_{g}), \xi_p(\boldsymbol{c}_{p}), \xi_m(\boldsymbol{c}_{m})]) \label{eqn:condition} \\
    & p_\theta\left(\boldsymbol{x}_{t-1} \mid \boldsymbol{x}_t, \boldsymbol{c}\right)=\mathcal{N}\left(\boldsymbol{x}_{t-1} ; \mu_\theta\left(\boldsymbol{x}_t, t, \boldsymbol{c}\right), \Sigma_\theta(t)\right)\label{eqn:denoise}\vspace{-5mm}
\end{align}
where each $\xi$ is some convolutional layers that transform the input to $\mathbb{R}^{m \times m \times d}$, $\varphi$ is another fusion layer,
% using convolutions that results in $\mathbb{R}^{m \times m \times d}$
$p_\theta$ is the probabilistic model that predicts the distribution of a less noisy image $\boldsymbol{x}_{t-1}$ from $\boldsymbol{x}_{t}$ with $t \in [0, T]$,
% is the timestamp
and $\mu_\theta$ and $\Sigma_\theta$ are the mean and variance of the normal distribution $\mathcal{N}$. We follow the same training losses in AnyText~\cite{tuo2023anytext} to train this diffusion model.

% NEED TO SAY SOMETHING ABOUT THE LOSSES USED HERE EVEN IF WE ARE JUST USING THE AnyNet training scheme.

\subsection{Inference Pipeline}
Our philosophy is to design a more streamlined training pipeline that is easily scalable to larger open-world datasets, while shifting additional steps to inference time to provide users with greater control and flexibility as needed.

\subsubsection{Main Generation Process}
The reversed denoising process takes the same set of inputs outlined in Section~\ref{sec:train_input}. However, unlike training where the glyph control $\boldsymbol{c}_g$ is extracted using the text segmentation model $\mathcal{M}$, it is now provided directly by the user. 

On the user front-end, the required inputs include the original image $\boldsymbol{I}$ with a short caption $\boldsymbol{c}_{l}$, the desired text $t$, the font $f$ (which can be uploaded as a font file), and the polygon points $\boldsymbol{p}$ selected on $\boldsymbol{I}$ to define the region where the text will be rendered.
To streamline the process, the pipeline automatically converts polygon points $\boldsymbol{p}$ into the position control $\boldsymbol{c}_{p}$, generates the masked image $\boldsymbol{c}_{m}$, and converts text $t$ into the font-aware glyph control $\boldsymbol{c}_{g}$. 
% the following inputs are then automatically prepared in the backend for the user:
% \begin{itemize}  
%     \item The position control $\boldsymbol{c}_{p}$ drawn from the user-specified polygon points $\boldsymbol{p}$. 
%     \item The glyph control generated by rendering the text $t$ in the selected font $f$ on a black canvas of size $n \times n$, with the text in white. A perspective transformation is then applied to automatically fit the rendered glyph onto the position $\boldsymbol{c}_{p}$, same as the ones used during training.
%     \item The masked image $\boldsymbol{c}_{m}$ drawn by masking out regions in $\boldsymbol{I}$ within $\boldsymbol{p}$.
%     % , i.e., $\boldsymbol{c}_{m} \in [0, 255]\ \text{where } \boldsymbol{c}_{m} = \boldsymbol{I} \ \text{but with } \boldsymbol{I} \mid_{\text{pixels within } \boldsymbol{p}} = 127.5.$.
% \end{itemize}
Users are allowed to type multiple lines of text, possibly in different languages, fonts, or orientations, in a single $\boldsymbol{c}_{p}$, $\boldsymbol{c}_{g}$, and $\boldsymbol{c}_{m}$.

Finally, the reverse denoising process is run over $t$ timesteps following the same Equations~\ref{eqn:condition}-\ref{eqn:denoise} to generate the output image $\boldsymbol{x}_{0}$, whose region within the polygon mask $\boldsymbol{c}_{p}$ is will be blended into the original image $\boldsymbol{I}$ using normal seamless cloning or other blending algorithms. This completes the generation process, and the next two subsections illustrate optional steps that can be applied as needed.

\begin{figure*}[!t]
    \centering
    \includegraphics[width=0.8\linewidth]{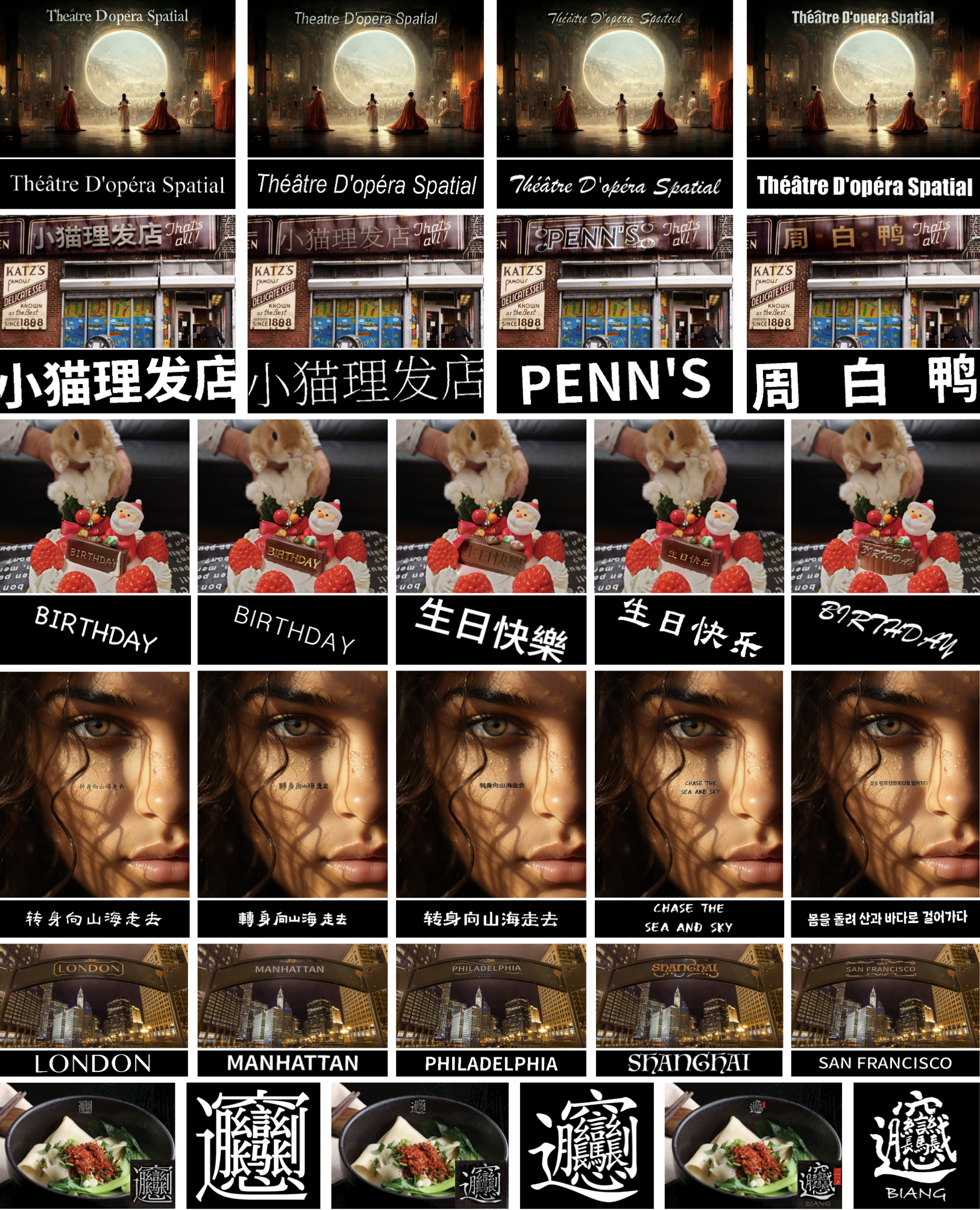}
    \caption{Continuation of Figure~\ref{fig:top}. Examples of real-world and AI-generated images with text generated by ControlText in various fonts and languages. Each row presents both the rendered images and the textual part of their glyph controls. We also try the most complex Chinese character, ``biang", in the bottom row, accompanied by a zoomed-in view of the rendered character. ControlText effectively renders text with realistic integration into backgrounds while maintaining correct letters and characters in their user specified fonts.}
    \label{fig:main}
    % \vspace{-1cm}
\end{figure*}

\subsubsection{Inpainting Before Editing}
When editing text in an image, the mask $\boldsymbol{c}_m$ must encompass all the old text in the background. However, this mask could be larger than the size of the new text $t$ in the new font $f$, particularly when a narrower font is selected. Larger masks may introduce additional text rendered in the output image not specified in the glyph control $\boldsymbol{c}_g$.
% Meanwhile, because the training process described in Section~\ref{sec:prepare_glyph} utilizes an OCR-based filtering, $\boldsymbol{I}$ may contain additional text elements not included in the final glyph control $\boldsymbol{c}_g$ during training, which the text segmentation model $\mathcal{M}$ does not segment well. As a result, masks larger than the new text in new font at inference, especially with out-of-domain user images $\boldsymbol{I}$, can introduce additional text in the output image $\boldsymbol{x}_0$ within the mask region but not specified in the glyph control $\boldsymbol{c}_g$.
To address this challenge, we minimize the mask size to be just large enough to fit the text $t$ in the new font $f$. Following recommendations in \cite{li2024first}, we utilize an off-the-shelf inpainting model~\cite{razzhigaev2023kandinsky} to erase the original text. After inpainting, a new polygon $\hat{\boldsymbol{p}}$ is automatically tightened
from $\boldsymbol{p}$ to match the new text.
% , ensuring it is just large enough to contain the new text $t$ in the new font $f$. Following the recommendations in \cite{li2024first}, we utilize an off-the-shelf inpainting model~\cite{razzhigaev2023kandinsky} to erase the original text from the user-defined polygon region $\boldsymbol{p}$. Once inpainting is complete, a new polygon $\hat{\boldsymbol{p}}$ is automatically tightened to match the size of the new text $t$ in the font $f$. The tighter polygon $\hat{\boldsymbol{p}}$ and the corresponding mask $\hat{\boldsymbol{c}}_m$ will be sent to the diffusion model, enabling cleaner output images that avoid unwanted text not mentioned by the user.

\subsubsection{Small Textual Regions}
Handling smaller text remains a challenge~\cite{liu2024glyph, paliwal2024customtext}, as the diffusion process operates in the embedding space with potential information loss. To address this, we simply zoom into the text region specified by the user and interpolate it to the input size of the diffusion model. Finally, we blend the generated region with the original image $\boldsymbol{I}$. Figure~\ref{fig:main} includes some examples of small text rendered with high quality, demonstrating effective performance without the need for more complex algorithms or datasets.

\subsection{Evaluation Metrics}
\subsubsection{Evaluating Text}
We adopt the same evaluation metrics from AnyText~\cite{tuo2023anytext} to ensure that the generated text remains recognizable regardless of the font. Specifically, we utilize Sentence Accuracy (SenACC) and Node similarity with Edit Distance (NED) derived from OCR to assess text recognizability. We also employ Fréchet Inception Distance (FID) to evaluate the overall image quality.

\subsubsection{Evaluating Fuzzy Fonts}
Evaluating the accuracy of fonts in visual text remains an open question, as ground-truth font labels are typically unavailable in large-scale, real-world datasets. It is also the case that many fonts appear visually similar, making distinctions among them practically meaningless.\textbf{ These challenges highlight the need for a new evaluation metric that can handle fuzzy fonts in an open-world scenario.}
To address this, we introduce a novel evaluation framework leveraging a pre-trained font classification model $\mathcal{F}$. Specifically, we use the Google Font Classifier by Storia-AI~\cite{fontclassify2025}, an open-source model trained on $c = 3474$ fonts on both real and AI-generated images. Due to the large value of $c$, the classifier's embedding space is expected to provide meaningful representations, \textit{even if the model may have never encountered the evaluated font before.} For example, \textbf{two fonts that look similar should have similar embeddings in this pretrained font classification model}, and vice versa. Therefore, we propose two metrics $\boldsymbol{l_2@k}$ and $\boldsymbol{\cos@k}$ to evaluate font fidelity in any generated images with text of any fonts.

\begin{itemize}
    \item \textbf{Step 1 Embedding Extraction:} Both the input glyph $\boldsymbol{c}_g$ and the output image $\boldsymbol{x}_0$ are forwarded through the font classification model $\mathcal{F}$ to obtain their last-layer probabilities $\boldsymbol{p}_g, \boldsymbol{p}_x \in \mathbb{R}^{c}$, respectively, where $c$ is the number of labels $\mathcal{F}$ is pretrained on. Optionally, text regions in $\boldsymbol{x}_0$ can be first isolated using a text segmentation model $\mathcal{M}$, eliminating the influence of color and background.
    \item \textbf{Step 2 Distance Calculation:} We retain only the top $k$ largest values in $\boldsymbol{p}_g$ and $\boldsymbol{p}_x$, zeroing out the others, to ensure that the distance calculation focuses on the most likely $k$ labels. It helps reduce disturbances from the accumulation of remaining insignificant values. The metric $l_2@k$ and $\cos@k$ then compute the $l_2$-distance and $\cos$-distance between them.

\end{itemize}
\section{Experimental Results}
\begin{table*}[!t]
\centering
\caption{Results on AnyText\textendash benchmark. ``AnyText with font\textendash aware glyphs'' denotes AnyText~\cite{tuo2023anytext} that directly adopts our glyph controls without fine-tuning as an ablation. ControlText consistently preserves detailed font information while maintaining strong text accuracy.}
\label{tab:results}
\small
\resizebox{\textwidth}{!}{
\begin{tabular}{l|cc|cccc|cccc}
\hline
\multicolumn{11}{c}{\textbf{English}}\\
\hline
\multirow{2}{*}{Method} &
\multicolumn{2}{c|}{\textbf{Text Accuracy} $\uparrow$} &
\multicolumn{8}{c}{\textbf{Fuzzy Font Accuracy (distance) $\downarrow$}}\\
\cline{2-11}
& SenACC & NED &
$l_{2}@5$ & $l_{2}@20$ & $l_{2}@50$ & $l_{2}@\text{full}$ &
$\cos@5$ & $\cos@20$ & $\cos@50$ & $\cos@{\text{full}}$ \\
\hline
ControlText (ours)              & \textbf{0.8345} & \textbf{0.9537} & \textbf{0.3431} & \textbf{0.3387} & \textbf{0.3382} & \textbf{0.3381} & \textbf{0.2710} & \textbf{0.2523} & \textbf{0.2504} & \textbf{0.2500}\\
AnyText~\cite{tuo2023anytext}   & 0.8315 & 0.9518 & 0.4654 & 0.4628 & 0.4623 & 0.4622 & 0.4125 & 0.3974 & 0.3954 & 0.3948\\
AnyText w/ glyphs               & 0.5524 & 0.8387 & 0.4738 & 0.4709 & 0.4705 & 0.4703 & 0.4261 & 0.4096 & 0.4077 & 0.4070\\
TextDiffuser~\cite{chen2024textdiffuser} & 0.5966 & 0.8236 & 0.6293 & 0.6280 & 0.6278 & 0.6277 & 0.5536 & 0.5448 & 0.5436 & 0.5432\\
GlyphControl~\cite{yang2023glyphcontrolglyphconditionalcontrol}  & 0.4098 & 0.7414 & 0.6046 & 0.6031 & 0.6029 & 0.6028 & 0.5423 & 0.5321 & 0.5307 & 0.5302\\
ControlNet~\cite{zhang2023adding}    & 0.5837 & 0.8305 & 0.6853 & 0.6839 & 0.6837 & 0.6837 & 0.5896 & 0.5812 & 0.5802 & 0.5798\\
\hline
\multicolumn{11}{c}{\textbf{Chinese}}\\
\hline
\multirow{2}{*}{Method} &
\multicolumn{2}{c|}{\textbf{Text Accuracy} $\uparrow$} &
\multicolumn{8}{c}{\textbf{Fuzzy Font Accuracy (distance) $\downarrow$}}\\
\cline{2-11}
& SenACC & NED &
$l_{2}@5$ & $l_{2}@20$ & $l_{2}@50$ & $l_{2}@\text{full}$ &
$\cos@5$ & $\cos@20$ & $\cos@50$ & $\cos@{\text{full}}$ \\
\hline
ControlText (ours)              & 0.7867 & 0.9276 & \textbf{0.3561} & \textbf{0.3508} & \textbf{0.3499} & \textbf{0.3497} & \textbf{0.3295} & \textbf{0.3015} & \textbf{0.2975} & \textbf{0.2964}\\
AnyText~\cite{tuo2023anytext}   & \textbf{0.8591} & \textbf{0.9515} & 0.4632 & 0.4593 & 0.4585 & 0.4583 & 0.4743 & 0.4488 & 0.4444 & 0.4431\\
AnyText w/ glyphs               & 0.5578 & 0.8120 & 0.4683 & 0.4646 & 0.4638 & 0.4636 & 0.4808 & 0.4553 & 0.4512 & 0.4490\\
TextDiffuser~\cite{chen2024textdiffuser} & 0.0611 & 0.2816 & 0.6773 & 0.6761 & 0.6757 & 0.6756 & 0.6638 & 0.6528 & 0.6509 & 0.6502\\
GlyphControl~\cite{yang2023glyphcontrolglyphconditionalcontrol}  & 0.0377 & 0.2338 & 0.7298 & 0.7285 & 0.7282 & 0.7281 & 0.6995 & 0.6891 & 0.6873 & 0.6865\\
ControlNet~\cite{zhang2023adding}    & 0.3500 & 0.6393 & 0.7609 & 0.7594 & 0.7591 & 0.7590 & 0.7123 & 0.7018 & 0.7001 & 0.6994\\
\hline
\end{tabular}}
\end{table*}

\begin{table*}[!t]
\centering
\caption{Fuzzy font accuracy on held-out Kannada and Korean languages in zero-shot settings. Lower values indicate better font similarity preservation. ControlText consistently outperforms AnyText with better generalization to completedly unseen languages, without explicitly being trained on them.}
\label{tab:fuzzy_subset}
\small
\begin{tabular}{l|cccc|cccc}
\hline
\multirow{2}{*}{Method} &
\multicolumn{8}{c}{Fuzzy Font Accuracy (distance) $\downarrow$} \\
\cline{2-9}
& $l_{2}@5$ & $l_{2}@20$ & $l_{2}@50$ & $l_{2}@\text{full}$ 
& $\cos@5$ & $\cos@20$ & $\cos@50$ & $\cos@{\text{full}}$ \\
\hline
ControlText (ours)        & \textbf{0.4564} & \textbf{0.6090} & \textbf{0.4507} & \textbf{0.6075} & \textbf{0.4503} & \textbf{0.6074} & \textbf{0.4503} & \textbf{0.6073} \\
AnyText~\cite{tuo2023anytext} & 0.4587 & 0.6209 & 0.4526 & 0.6193 & 0.4522 & 0.6192 & 0.4521 & 0.6192 \\
\hline
\end{tabular}
\end{table*}

\subsection{Experimental Setups}
We finetune the ControlNet~\cite{zhang2023adding} model, with a size of around $1$B pretrained by AnyText~\cite{tuo2023anytext} and several convolutional encoders to process input conditions, for 10 epochs using 4 NVIDIA V100 GPUs, each with 32 GB memory. 
% Due to limitations in our computational resources, each epoch in total requires around 380 GPU hours. 
We use a batch size of 6, a learning rate of $2 \times 10^{-5}$, and focus solely on inpainting masked images.
The dataset is curated from AnyWord-3M~\cite{tuo2023anytext} but with our font-aware glyph. Each RGB image of size $512$ by $512$ has at most $5$ lines of text. The dataset comprises approximately 1.39 million images in English, 1.6 million images in Chinese, and 10 thousand images in other languages. Following this, we continue training the model for another 2 epochs, turning on the textual perception loss introduced in~\citet{tuo2023anytext}. We use AnyText-benchmark~\cite{tuo2023anytext} with 1000 test images in English and Chinese to show quantitative results.

\subsection{Visual Results}
Figures~\ref{fig:top} and~\ref{fig:main} showcase open-world images generated by our model. We always follow the text editing pipeline to either modify existing text or render new text. The original images $\boldsymbol{I}$ used in our experiment include both real~\cite{pixta_photo_93233725, unsplash, businessinsider_takeout, cnn_nyc_restaurants, peterpom211_x_post, tripadvisor_chicago_dark_side, nipic_image} and AI-generated~\cite{theatre_dopera_spatial, midjourney_prompts} examples. ControlText demonstrates high-fidelity text rendering, accurately preserving both the text and the font styles. It automatically render text in either flat formats or with depth and color effects based on the background, such as outward-engraved text on a shabby storefront sign on the street, a metallic board on wall, a chocolate bar, or with neon light effects at night. 

We present images in multiple languages: English, French (zero-shot), traditional and simplified Chinese (including \textit{the} most complex character \textit{``biang"}), Japanese (including Kaomoji), and Korean, rendered in either single or multi-line formats. Additionally, we incorporate various font styles, including novel designer fonts sourced from the web~\cite{apple_fonts, fonts_net_cn}.

\subsection{Quantitative Results}
Table~\ref{tab:results} presents the quantitative results evaluated on the AnyText benchmark~\cite{tuo2023anytext}, along with our proposed metrics $l_2@k$ and $\cos@k$ with $k = 5, 20, 50$, and the full logits, i.e., $c = 3474$ in the pretrained font classification model to assess font fidelity. ControlText generates glyphs via a segmentation model, which may yield occasional low-quality masks. Since users can provide high-quality glyphs in practice, these results serve as lower bounds. For fairness, we filter out low-quality masks with criterion from Section~\ref{sec:prepare_glyph}.

To further evaluate the cross-lingual generalization ability of our model, we conducted zero-shot inferences on Kannada and Korean scripts from the MLe2e dataset \cite{gomez2016finegrainedapproachscenetext}, previously unseen by the model. As shown in Table \ref{tab:fuzzy_subset}, ControlText consistently outperforms AnyText across all metrics. Specifically, ControlText achieves lower average cosine and l2 distances across top-5, top-20, top-50, and full-set comparisons in font accuracy metrics, indicating better font-style consistency and generation quality. Meanwhile, these results reinforce that ControlText generalizes better to underrepresented scripts such as Kannada and Korean, even without explicit training on these languages.

While ControlText shows some differences compared to AnyText in  SenACC and NED on Chinese characters, it successfully maintains large gaps across metrics on English data and fuzzy font accuracy. Meanwhile, when using identical font-aware glyph controls in ControlText, AnyText experiences a substantial decrease in text accuracy with almost no improvement in font accuracy, as shown in the row marked ``AnyText-v1.1 Font Aware" in Table~\ref{tab:results}. This demonstrates ControlText’s superior ability to handle diverse and nuanced font variations without requiring fine-tuning for each font. 

\section{Discussion}
This work presents a simple and scalable proof-of-concept for multilingual visual text rendering with user-controllable fonts in the open world. We summarize our key findings as follows:

\paragraph{Font controls require no font label annotations} 
A text segmentation model can capture nuanced font information in pixel space without requiring font label annotations in the dataset, enabling zero-shot generation on unseen languages and fonts (as evidenced by generated images with Japanese in Figure \ref{fig:top}), as well as scalable training on web-scale image datasets as long as they contain text.

\paragraph{Evaluating ambiguous fonts in the open world} 
Fuzzy font accuracy can be measured in the embedding space of a pretrained font classification model, utilizing our proposed metrics $l_2@k$ and $\cos@k$.

\paragraph{Supporting user-driven design flexibility} 
Random perturbations can be applied to segmented glyphs. While this won’t affect the rendered text quality, it accounts for users not precisely aligning text to best locations and prevents models from rigidly replicating the pixel locations in glyphs.

\paragraph{Working with foundation models} 
With limited computational resources, we can still copilot foundational image generation models to perform localized text and font editing.
\\

Future work will focus on improving data efficiency in the training pipeline, particularly for fonts in low-resource languages. We also plan to explore reinforcement learning with text and font verification signals in multimodal transformers. Additionally, we aim to enable more advanced artistic style control of text based on user prompts, extending beyond font attributes to include interactions with diverse background content.
\section{Limitations}
Our model is based on ControlNet~\cite{zhang2023adding} with a CLIP text embedding model~\cite{radford2021learning}, although modified by AnyText~\cite{tuo2023anytext} to incorporate glyph line information. However, the CLIP-text encoder has relatively limited language understanding capabilities compared to state-of-the-art foundation models. Unlike text itself, this limitation affects the model's ability to accurately render complex artistic visual features or backgrounds, which users might specify in their input prompts, such as asking the text to appear like clouds or flames, that go beyond merely the font information.

Additionally, due to limited training resources, our experiments were conducted using a smaller diffusion model as a proof-of-concept compared to commercial ones. Each epoch requires approximately 380 GPU hours on NVIDIA V100 GPUs with 32 GB of memory, but we anticipate significantly improved efficiency on newer hardware and with a larger memory. This constraint may result in suboptimal inpainting of background regions within the text area, as well as instability in the quality of rendered text. This also made it difficult to try other base models for comparison. Moreover, the limited training resources also made it difficult to conduct more ablation studies, which could have provided more nuanced insights into the architecture's effectiveness. The users also have limited controls of background pixels behind the text.  

Some sacrifice in text quality is observed for non-Latin languages on the AnyText-Benchmark in exchange for improved font controllability.

The embedding layers of the glyph controls can also lead to reduced text quality, especially when the text in a font is very small, thin, or excessively long. In such cases, fine details of the font information in the glyphs may be lost. 

As with all other text-to-image algorithms that rely on diffusion models, our approach requires a certain number of denoising steps to generate a single image at inference. End-to-end transformer-based models~\cite{xie2024show} may improve the time efficiency of the generation process.

% The entire pipeline also requires a user-friendly front-end interface to allow users to create glyph control images efficiently, especially working with more complex text layouts or curved typography effects. However, as long as users specify the input glyph controls, our ControlNet is already prepared to follow the pixel-level information to render the image with the required text and font. 

\section{Ethical Impact}
This work is intended solely for academic research purposes. While our algorithm allows users to generate images with customized text, there is a potential risk of misuse for producing harmful or hateful content or misinformation. However, we do not identify any additional ethical concerns compared to existing research on visual text rendering.

\section{Acknowledgment}
This work was supported by the National Science Foundation (NSF) grant CCF-2112665 (TILOS).

\begingroup
\sloppy
\bibliography{custom}
\endgroup

% \appendix

% \section{Example Appendix}
% \label{sec:appendix}

% This is an appendix.

\end{document}